\documentclass[journal]{IEEEtran}
\usepackage {multirow}
\usepackage{color}
\usepackage[colorlinks,linkcolor=black]{hyperref}
\hypersetup{urlcolor=black}
\usepackage{amsmath}
\ifCLASSINFOpdf
   \usepackage[pdftex]{graphicx}
   \graphicspath{{../pdf/}{../jpeg/}}
   \DeclareGraphicsExtensions{.pdf,.jpeg,.png}
\else
   \usepackage[dvips]{graphicx}
   \graphicspath{{../eps/}}
   \DeclareGraphicsExtensions{.eps}
\fi
\begin{document}
\title{Infrared and Visible Image Fusion via Interactive Compensatory Attention Adversarial Learning}


\author{Zhishe~Wang,\textit{~Member,~IEEE},~Wenyu~Shao,~Yanlin~Chen,~Jiawei~Xu,~Xiaoqin~Zhang,\textit{~Member,~IEEE}% <-this % stops a space
\thanks{This work is supported in part by Fundamental Research Program of Shanxi Province under Grant 201901D111260, in part by the Open Foundation of Shanxi Key Laboratory of Signal Capturing \& Processing under Grant ISPT2020-4. (\textsl{Corresponding author: Jiawei~Xu}).}% <-this % stops a space
\thanks{Zhishe~Wang, Wenyu~Shao and Yanlin~Chen are with School of Applied Science, Taiyuan University of Science and Technology, Taiyuan, 030024, China. (e-mail: wangzs@tyust.edu.cn; wyshaotyust@163.com; chentyust@163.com)

Jiawei~Xu and Xiaoqin~Zhang are with the Institute of Big Data and Information Technology, Wenzhou University, and also with the College of Computer Science and Artificial Intelligence, Wenzhou University, Wenzhou, 303205, China. (e-mail: jxulincoln@gmail.com; zhangxiaoqinnan@gmail.com)}

% <-this % stops a space
}
\maketitle

\begin{abstract}
The existing generative adversarial fusion methods generally concatenate source images and extract local features through convolution operation, without considering their global characteristics, which tends to produce an unbalanced result and is biased towards the infrared image or visible image. Toward this end, we propose a novel end-to-end mode based on generative adversarial training to achieve better fusion balance, termed as \textit{interactive compensatory attention fusion network} (ICAFusion). In particular, in the generator, we construct a multi-level encoder-decoder network with a triple path, and adopt infrared and visible paths to provide additional intensity and gradient information. Moreover, we develop interactive and compensatory attention modules to communicate their pathwise information, and model their long-range dependencies to generate attention maps, which can more focus on infrared target perception and visible detail characterization, and further increase the representation power for feature extraction and feature reconstruction. In addition, dual discriminators are designed to identify the similar distribution between fused result and source images, and the generator is optimized to produce a more balanced result. Extensive experiments illustrate that our ICAFusion obtains superior fusion performance and better generalization ability, which precedes other advanced methods in the subjective visual description and objective metric evaluation. Our codes will be public at \url{https://github.com/Zhishe-Wang/ICAFusion}.

\end{abstract}


% \fi
\IEEEpeerreviewmaketitle

\section{Introduction}
\IEEEPARstart{I}{nfrared} sensors can perceive heat source target characteristics by receiving thermal radiation, and work at different times or any weather conditions, however, the obtained images often represent high-brightness targets by pixel intensity, but lack structural textures. On the contrary, visible sensors can characterize rich scene and texture details through light reflection, but fail to identify significant targets, and are sensitive to light conditions, espically in low illumination environments. Since these two kinds of sensors have strong complementarity in imaging conditions and imaging mechanisms, image fusion technology can effectively overcome their own shortcomings and adequately fulfill their respective advantages to achieve a more informative image with prominent target perception and abundant detail characterization, which can benefit other subsequent tasks, such as RGBT tracking [1], RGB-D salient object detection [2] and multi-spectral pedestrian re-recognition [3], etc.

\begin{figure*}[!t]
	\centering
	\includegraphics[width=1\textwidth]{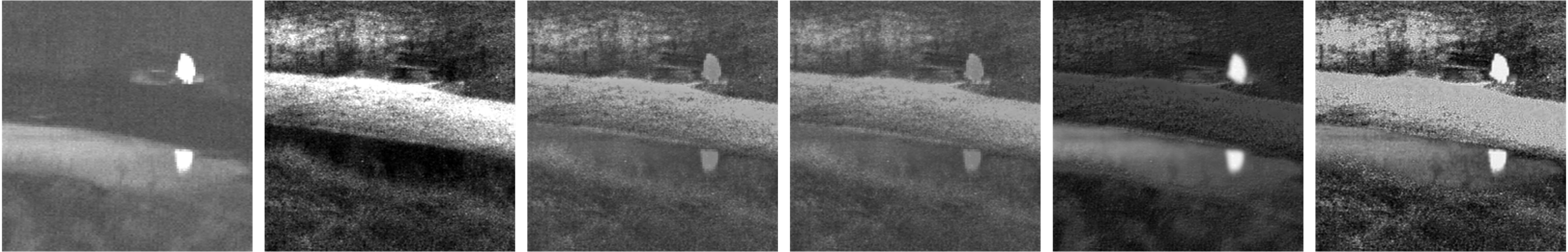}
	% where an .eps filename suffix will be assumed under latex, 
	% and a .pdf suffix will be assumed for pdflatex; or what has been declared
	% via \DeclareGraphicsExtensions.
	\caption{The contrastive schematic illustration of our proposed ICAFusion. The left two images are source images, and others are the fusion images obtained by MDLatLRR [7], DenseFuse [15], FusionGAN [19] and our ICAFusion, respectively.}
	\label{Fig1}
\end{figure*}

The existing traditional fusion methods usually employed a fixed mathematical model based on prior knowledge of target characteristics and imaging mechanism to extract features, designed an appropriate strategy to combine them, and then reconstructed the final fusion image through the corresponding inverse operations. The representative methods are multi-scale transformation [4, 5], sparse representation [6, 7], saliency-based [8], subspace-based [9] and mimicry fusion [10] and others [11, 12]. Typically, Li \textit{et al.} [7] presented MDLatLRR where source images were decomposed by multi-level latent low-rank representation into base and detail parts, and proposed average and nuclear-norm as the corresponding fusion strategies. The learnable low-rank representation can potentially increase the extraction ability of salient features, and further achieve better fusion performance, but its computational efficiency is very low. In fact, due to different imaging mechanisms, infrared images represent target characteristics by pixel intensity, while visible images characterize scene textures by edges and gradient. The traditional fusion methods fail to consider their inherent distinctiveness, and employ a uniform mathematic model to indiscriminately extract image features. However, the proposed mathematic model is only sensitive to a certain feature, and may not be suitable for other features, which inevitably leads to low fusion performance and poor visual effect in some cases. In addition, the corresponding fusion strategy is manually designed and increasingly complicated, which severely hinders the practical application of image fusion.

Recently, due to the improvement of machine learning and hardware devices, deep learning has greatly promoted the fast development of image fusion [13]. The convolutional neural network (CNN) based methods [14-17] generally introduced the encoder-decoder network framework for feature extraction and feature reconstruction. For example, Li \textit{et al.} [15] proposed DenseFuse in which the intermediate features were resued by employing a densely connected block to enhance feature representation power, and their fusion network was easy to be steadily trained because MS-COCO [18] dataset was adopted. However, these methods are non-end-to-end model, and fusion strategy still need to be manually designed. To address this drawback, the generative adversarial network (GAN) based methods [19-21] were developed to transform image fusion into an adversarial game. Typically, Ma \textit{et al.} [19] exploited FusionGAN where the discriminator continuously optimized the generator by adversarial training to achieve the similar distribution between fused result and source images. Although the GAN-based methods have achieved remarkable effects, some non-negligible issues need to be further overcomed. On the one hand, these methods concatenate source images as the input image, and only rely on a discriminator to perform the adversarial training, which leads to insuﬀicient local details and blurred target edges in the fusion image. On the other hand, these methods only depend on the convolutional operations to extract local features, but fail to consider their global dependencies, which cannot effectively maintain infrared targets and visible details simultaneously.

To overcome the above-mentioned issues, we develop an interactive compensatory attention fusion network for infrared and visible images, namely ICAFusion. Firstly, we propose a novel end-to-end fusion mode based on the wasserstein generative adversarial network [22] that does not require human participation, which overcomes the limitation of a hand-designed fusion strategy. Secondly, we construct a multi-level encoder-decoder network in the generator, which consists of a triple path, \textit{i.e.,} infrared, visible and their concatenating path. The infrared and visbile paths are communicated to provide intensity and gradient information for the concatenated path, which can retain more infrared pixel intensity and visible gradient information for the subsequent processing. Thirdly, we develop interactive and compensatory attention modules, which cascade the channel and spatial models, to model the long-range dependences and transfer features for the triple path. The interactive attention modules are applied to interact features for the encoder, while the compensatory attention modules are used to compensate features for the decoder. The obtained attention maps mix up with the local and global characteristics to achieve high performance feature extraction and feature reconstruction. Finally, we design dual discriminators, \textit{i.e.,} the Discriminator-IR and Discriminator-VIS, to identify the similar distribution between fused result and source images, and optimize the generator to produce a more balanced fused result.

To intuitively demonstrate our fusion performance, a contrastive schematic illustration is presented in Fig.1. Very obviously, the traditional MDLatLRR [7] and CNN-based method, \textit{i.e.,} DenseFuse [15], tend to retain more visible detail information, but lose the brightness of infrared targets. On the contrary, the GAN-based method, \textit{i.e.,} FusionGAN [19], is inclined to contain high-brightness infrared target information, but target edges are blurred and visible texture details are seriously missing. In contrast, our ICAFusion not only retains infrared typical targets but also reserves abundant visible details, and achieves better visual perception with higher image contrast.

Our main contributions can be summarized as four aspects:

$ \bullet $ We construct a multi-level encoder-decoder network with a triple path in the generator. The individual infrared and visible paths provide additional intensity and gradient information for the concatenating path under feature interaction and feature compensation, which can preserve more significant infrared targets and abundant visible details in the fusion image.

$ \bullet $ We develop interactive and compensatory attention modules to communicate their pathwise information for the triple path, and model the global features from the channel and spatial dimensions, which can increase feature representation power to more place emphasis on infrared target perception and visible detail characterization.

$ \bullet $ We design dual discriminators to supervise and optimize the generator. The Discriminator-IR and Discriminator-VIS are used to more evenly identify the similar distribution between fused result and source images. The desired generator can produce a more balanced fused result with more similar pixel distribution and finer texture details from source images.

$ \bullet $ We propose an end-to-end wasserstein generative adversarial network for infrared and visible image fusion. Extensive experiments indicate that our ICAFusion precedes other representative state-of-the-art fusion methods in the subjective visual description and objective metric evaluation.

The rest of this paper is organized as follows. Section II presents the development of CNN-based and GAN-based fusion methods. Section III clarifies the problem formulation and describes the network framework, attention modules and loss function. The related experiments and conclusion are discussed in Section IV and V, respectively.

\section{Related work}
In this section, we comprehensively review the representative CNN-based and GAN-based fusion methods, and further discuss their superiority and drawbacks.

\begin{figure*}[!t]
	\centering
	\includegraphics[width=1\textwidth]{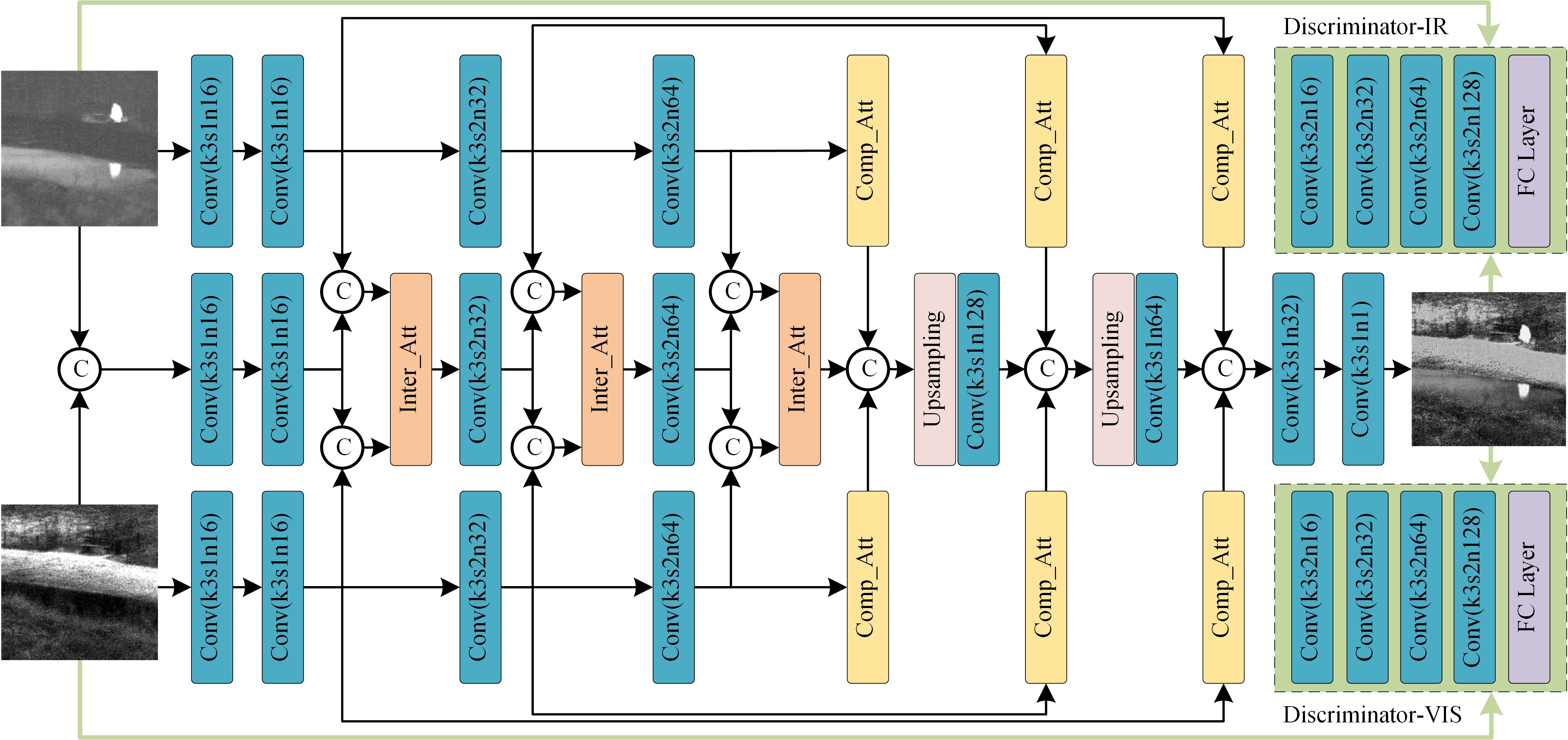}
	% where an .eps filename suffix will be assumed under latex, 
	% and a .pdf suffix will be assumed for pdflatex; or what has been declared
	% via \DeclareGraphicsExtensions.
	\caption{The principle of our ICAFusion with a triple path, which includes a generator and dual discriminators, $ i.e., $ Discriminator-IR and Discriminator-VIS. Inter\_Att and Comp\_Att denote interactive and compensatory attention modules, respectively. \textcircled{c} represents concatenation operation.}
	\label{Fig2}
\end{figure*}

\subsection{CNN-based fusion methods}
Compared with the traditional fusion methods, the convolutional neural network employs more filter banks to automatically extract features from the training dataset, which can reduce the imperfection of the hand-craft feature extraction model, and further improve image fusion performance. For example, Jian \textit{et al.} [14] proposed the modified residual dense network to decompose deep features, and applied a visual saliency mechanism to generate their corresponding decision maps to guide feature combinations. However, the proposed network is simple, and not especially training for fusion task. Li \textit{et al.} [15] presented DenseFuse where a densely connected block was applied to reemploy the intermediate features, average and $L_{1}$ norm were adopted as fusion strategies. Luo \textit{et al.} [16] exploited a multi-branch network with contrastive constraints, and designed a general fusion rule based on the disentangled representation. Zhang \textit{et al.} [17] introduced a general training network with a simple average rule for the multitask image fusion. These methods rely entirely on convolutional operations to extract local features, but ignore their long-rang dependencies and inevitably lose the important global information to some extent.

In order to exploit the local and global features to achieve better representational capacity, Jian \textit{et al.} [23] introduced SEDRFuse in which a symmetric network framework was proposed, and the spatial attention fusion strategy was designed. Li \textit{et al.} [24] presented NestFuse where a decoder network based on nest connections was designed for better feature reconstruction, spatial-wise and channel-wise attention models were proposed as fusion strategies. Wang \textit{et al.} [25] developed Res2Fusion in which two multiple receptive field aggregation blocks were proposed to generate multi-level features, and fusion strategies based on channel and spatial nonlocal attention models were designed. Subsequently, Wang \textit{et al.} [26] introduced UNFusion where a unified multi-scale dense network was designed, and $L_{p}$ normalized attention models were proposed to establish the long-range dependencies of local features. Although these methods have achieved supernormal results, their attention fusion strategies are manually designed and not learnable.

To overcome the limitations of hand-designed feature fusion, Long \textit{et al.} [27] exploited an unsupervised aggregated residual dense network for infrared and visible image fusion, which designed pixel-wise and feature-wise loss functions to supervise the network. Li \textit{et al.} [28] employed a two-stage training mode, namely RFN-Nest, which first trained the encoder-decoder network, and then trained the residual fusion module. Furthermore, for the multitask image fusion, Zhao \textit{et al.} [29] designed a novel universal framework to learn specific and general features, and proposed a realm activation mechanism to facilitate high generalization of across-realm. Xu \textit{et al.} [30] proposed a novel unified and unsupervised network to solve multiple fusion problems, which applied the information preservation degrees to constrain the loss function by measuring the importance of corresponding source images. Zhang \textit{et al.} [31] presented PMGI where the gradient and intensity paths were performed to realize different image fusion tasks. These methods are end-to-end mode without designing a hand-designed fusion strategy. However, they focus on the design of network structure and loss function, and still fail to model the global features, which inevitably cause the loss of some contextual information in the fusion image. 

\begin{figure*}[!t]
	\centering
	\includegraphics[width=1\textwidth]{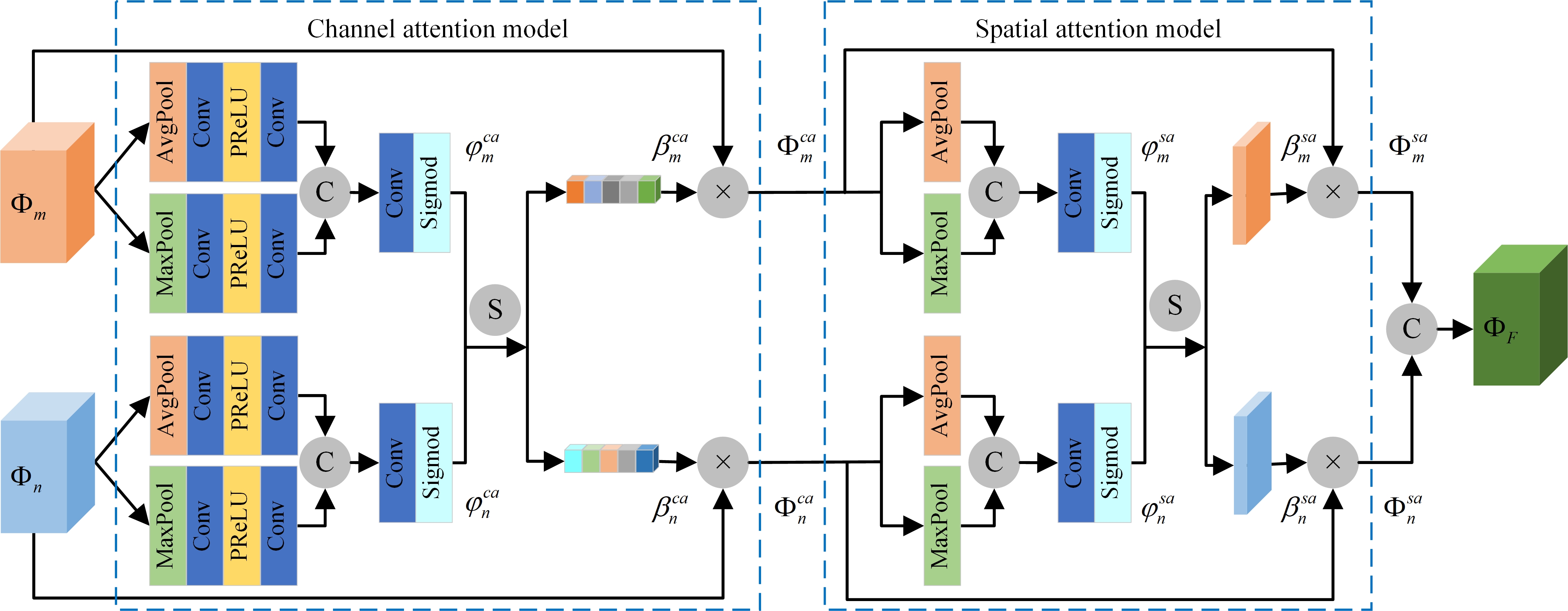}
	% where an .eps filename suffix will be assumed under latex, 
	% and a .pdf suffix will be assumed for pdflatex; or what has been declared
	% via \DeclareGraphicsExtensions.
	\caption{The network architecture of our interactive attention module, which cascades channel and spatial attention models. \textcircled{s} and \textcircled{$ \times $} denotes softmax and multiplication operations, respectively.}
	\label{Fig3}
\end{figure*}

\subsection{GAN-based fusion methods}

Different from the aforementioned methods, some reseachers translated fusion problem into a feature adversarial training. Typically, Ma \textit{et al.} presented FusionGAN [19] and its extended version [20] for image fusion tasks. Since their methods only use a discriminator, the obtained fused result is similar to an sharpened infrared image, and seriously lost the texture details of the visible image. To alleviate this problem, they specifically designed two discriminators to realize fusion balance, and exploited DDcGAN [32] to implement multi-resolution fusion tasks. In addition, Zhou \textit{et al.} [33] developed SDDGAN where an information quantity discrimination block was designed to supervise semantic information of source images under the framework of dual-discriminator generative adversarial network. Ma \textit{et al.} [34] translated image fusion into multi-classification constraints, namely GANMcC, which proposed two multi-classification discriminators to generate a more balanced result. These methods concatenate infrared and visible images as an input source, the fusion image maintains a limited balance, indicating that the result is inclined to a sharpen infrared image, and still lacks visible details.

In order to settle these issues, Li \textit{et al.} [35] employed a multi-grained attention network with two independent encoders, namely MgAN-Fuse, which integrated a channel attention model into multi-scale layers of the encoder, and then multi-grained attention maps were reconstructed a fused image by the decoder. Subsequently, they extended the attention mechanism into generator and discriminator, termed as AttentionFGAN [36], which designed two multi-scale attention networks to generate the respective attention maps of infrared and visible images, and were directly concatenated with source images for the fusion network to produce a fused result. These methods only adopt channel attention mechanism to enhance feature representation, but ignore its spatial characteristics. More importantly, the attention interaction and compensation are also not considered in feature encoding and decoding stages, which limits the fusion performance.

\section{Method}

\subsection{Problem Formulation}

For image fusion, the purpose of the generative adversarial network is to train the generator by fooling the discriminator, so that the generator can produce a more informative and better visual perceptive image. However, infrared and visible images have respective intrinsic distinctiveness, and their representative contents vary greatly under different imaging mechanisms. The infrared image retains high-brightness target characteristics in which the pixel intensity represents the histogram distribution of the target, while the visible image contains rich scene information in which the pixel difference, $ i.e.,$ edges and gradient characterize texture details of a scene. Rather than only concatenating infrared and visible images, we tend to solve their fusion problem from the essential characteristics of respective imaging. Therefore, we construct a multi-level encoder-decoder network framework with a triple path to extract the features, infrared and visible paths provide additional intensity and gradient information for the concatenating path, which can improve the representation ability for feature encoding and feature decoding. More specifically, we develop the interactive and compensatory attention module to communicate their pathwise information, and model their global features, which can refine features to more focus on infrared target perception and visible detail characterization. In addition, we design dual discriminators to identify the similar distribution between fused results and source images under the supervision of the specific loss function with pixel intensity and gradient variation constraints. The Discriminator-IR force the fusion image to distinguish the similar pixel intensity distribution from the infrared image, while the Discriminator-VIS force the fused result to identify the similar edges and gradient from the visible image. Each discriminator is used to preserve and enhance its corresponding modality features, and make the generator to produce a more balanced result.

\subsection{Network overview}
As shown in Fig.2, the proposed ICAFusion is based on the wasserstein generative adversarial network, which consists of a generator and dual discriminators. 

\textbf{Generator Architecture:} The generator includes the encoder part, fusion layer and decoder part. In the encoder part, a triple path, namely infrared, visible and their concatenating path, is proposed as input sources. We use four convolutional layers to extract multi-level features for the triple path, in which the third and fourth layers are a strided convolution with the factor of 2. The features of infrared and visible paths are respectively concatenated with that of the concatenating path, termed as $ {\Phi_m}$ and $ {\Phi _n}$, and then fed into an interactive attention module to produce their interactive attention maps, termed as $ {\Phi _{F}}$. After three level feature interactions, the final intractive attention maps are obtained. In the fusion layer, these final intractive attention maps are directly concatenated with the compensatory attention maps of infrared and visible paths to generate the fused attention maps. Subsequently, in the decoder part, we also use four convolutional layers to reconstruct features, where the first two layers are along with upsampling operation. The obtained output is concatenated with the corresponding compensatory attention maps of infrared and visible paths for subsequent reconstruction. In the end, we obtain the initial fusion image. All the layers use $ 3\times3 $ convolution kernels along with PReLU activation, except for the last layer with Tanh function.

\textbf{Discriminator Architecture:} The Discriminator-IR and Discriminator-VIS have the same network framework, which consist of four convolution layers and a fully connected layer. All the convolution layers are the strided operations with $3 \times 3 $ kernel size and LeakyRelu activation function. The stride is set to 2, and the corresponding filter banks are set to 16, 32, 64 and 128. During the training process, we input the initial fusion image $ {I _{f}}$, infrared image $ {I _{ir}}$ and visible image $ {I _{vis}}$ into the corresponding discriminator, which aim to distinguish $ {I _{f}}$ from $ {I _{ir}}$ and $ {I _{vis}}$. The Discriminator-IR force $ {I _{f}}$ to gradually preserve more and more infrared pixel intensity information, while the Discriminator-VIS force $ {I _{f}}$ to increasingly contain more and more visible detail information. When the adversarial game of the generator and dual discriminators reaches equilibrium, it indicates that the generator has fooled dual discriminators, and the desired fused result is obtained, which can maintain more similar infrared pixel intensity and finer visible texture details at the same time.

\subsection{Interactive and compensatory attention modules}
Inspired by CBAM [37], we redesign and construct interactive and compensatory attention modules to communicate the pathwise information and model the global features. The framework of the interactive attention module is shown in Fig.3. For the intermediate features $ {\Phi _m}$ and $ {\Phi _n}$$ \in {R^{H \times W \times C}} $, we first employ global average and maximum pooling operations to aggregate feature maps into channel descriptions, respectively. Both descriptions pass through two convolutional layers with $ 3 \times 3 $ kernel size and a PReLU activation layer, the output feature vectors are concatenated together, and forwarded to the convolutional layer and sigmoid activation layer. In short, after the channel attention model, we obtain their respective initial channel weighted coefficients $ \varphi _m^{ca} $ and $ \varphi _n^{ca} $ $ \in {R^{1 \times 1 \times C}} $, which are computed by Eq.1 and 2. 
\begin{equation}
	\begin{split}
	\varphi _m^{ca}(c) = \delta (Conv(Con[Conv(\sigma (Conv(AP({\Phi _m})))), \\
	{\rm{ }}Conv(\sigma (Conv(MP({\Phi _m}))))]))
    \end{split}
\end{equation}
\begin{equation}
\begin{split}	
	\varphi _n^{ca}(c) = \delta (Conv(Con[Conv(\sigma (Conv(AP({\Phi _n})))),\\ 
	{\rm{ }}Conv(\sigma (Conv(MP({\Phi _n}))))]))
\end{split}
\end{equation}
where $ Conv $ and $ Con $ represent the convolution and concatenation operations, $ AP( \cdot ) $ and $ MP( \cdot ) $ denote global average and maximum pooling operations, respectively. $ \sigma $ and $ \delta $ represent PReLU and sigmoid activation functions.

And then, we apply softmax operation to produce their final channel weighted coefficients, $ i.e.,$ $ \beta _m^{ca} $ and $ \beta _n^{ca} $, which are formulated by Eq.3 and 4.
\begin{equation}
	\beta _m^{ca}(c) = \frac{{\exp (\varphi _m^{ca}(c))}}{{\exp (\varphi _m^{ca}(c)) + \exp (\varphi _n^{ca}(c))}}
\end{equation}
\begin{equation}
	\beta _n^{ca}(c) = \frac{{\exp (\varphi _n^{ca}(c))}}{{\exp (\varphi _m^{ca}(c)) + \exp (\varphi _n^{ca}(c))}}
\end{equation}

We multiply the final channel weighted coefficients with their respective input features to obtain their corresponding channel attention maps, which are expressed by Eq.5 and 6. 
\begin{equation}
	\Phi _m^{ca}(i,j) = {\Phi _m}(i,j) \times \beta _m^{ca}(c)
\end{equation}
\begin{equation}
	\Phi _n^{ca}(i,j) = {\Phi _n}(i,j) \times \beta _n^{ca}(c)
\end{equation}

Subsequently, the corresponding channel attention maps are tooken as the input of the spatial attention model, and forwarded to the global average and maximum pooling layers. The output spatial feature maps are concatenated together, and fed into a convolutional layer and a sigmoid activation layer, we obtain their respective initial spatial weighted coefficients, which are computed by Eq.7 and 8. 
\begin{equation}
	\varphi _m^{sa}(i,j) = \delta (Conv(Con[AP(\Phi _m^{ca}),MP(\Phi _m^{ca})]))
\end{equation}
\begin{equation}
	\varphi _n^{sa}(i,j) = \delta (Conv(Con[AP(\Phi _n^{ca}),MP(\Phi _n^{ca})]))
\end{equation}

And then, we apply softmax operation to produce their final spatial weighted coefficients, $ i.e.,$ $ \beta _m^{sa} $ and $ \beta _n^{sa} $, which are formulated by Eq.9 and 10.
\begin{equation}
	\beta _m^{sa}(i,j) = \frac{{\exp (\varphi _m^{sa}(i,j))}}{{\exp (\varphi _m^{sa}(i,j)) + \exp (\varphi _n^{sa}(i,j))}}
\end{equation}
\begin{equation}
	\beta _n^{sa}(i,j) = \frac{{\exp (\varphi _n^{sa}(i,j))}}{{\exp (\varphi _m^{sa}(i,j)) + \exp (\varphi _n^{sa}(i,j))}}
\end{equation}

We multiply the final spatial weighted coefficients with their channel attention maps to produce their respective spatial attention maps, which are computed by Eq.11 and 12. 
\begin{equation}
	\Phi _m^{sa}(i,j) = {\Phi _m^{ca}(i,j) \times \beta _m^{sa}(i,j)}
\end{equation}
\begin{equation}
	\Phi _n^{sa}(i,j) = {\Phi _m^{ca}(i,j) \times \beta _n^{sa}(i,j)}
\end{equation}

Finally, we directly concatenate their corresponding spatial attention maps to produce the fused attention maps, which are expressed by Eq.13.
\begin{equation}
	\Phi _{F}(i,j) = Con[\Phi _m^{sa}(i,j), \Phi _n^{sa}(i,j)]
\end{equation}

Note that the compensatory attention module is equivalent to the upper part of the interactive attention module with only an intermediate feature input, and does not require the softmax operation. In other words, the features of infrared or visible image are in turn fed into the channel and spatial attention models to produce their respective attention maps, which are used to compensate information for feature reconstruction.

\subsection{Loss function}
In the proposed ICAFusion, we need to design the loss funcution of the generator and dual discriminators, respectively In the generator, the loss function consists of adversarial loss $ L_{adv} $ and content loss $ L_{con} $, which is expressed by Eq.14.
\begin{equation}
	{L_G} = {L_{adv}} + {L_{con}}
\end{equation}

Considering that infrared image represents target characteristics by pixel intensity, while visible image characterizes scene textures by edges and gradient. In this paper, we adopt frobenius norm and $ L_{1} $ norm to constrain the fused result with the similar pixel intensity and gradient variation of infrared and visible images, respectively. Therefore, the content loss function is expressed by Eq.15.
\begin{equation}
	{L_{con}} = \frac{1}{{HW}}(||{I_f}-{I_{ir}}||_F^2 + ||\nabla{I_f}-\nabla{I_{vis}}|{|_1})\
\end{equation}
where H and W represent the height and width of the source image, respectively. $ ||\cdot||_F $ and $ ||\cdot||_1 $ denote frobenius norm and $ L_{1} $ norm, $ \nabla $ indicates the gradient operator.

In the dual discriminators, the Discriminator-IR ($ D_{r} $) and Discriminator-VIS ($ D_{v} $) are designed to balance the authenticity of the fused result and source images, so that the generated result more tends to the real data distribution of source images. The adversarial loss function is expressed by Eq.16.
\begin{equation}
	{L_{adv}} = - \frac{1}{N}\sum\limits_{n = 1}^N {[{D_{r}}({I^n_f})]} - \frac{1}{N}\sum\limits_{n = 1}^N {[{D_{v}}({I^n_f})]}
\end{equation}
Meanwhile, the respective loss function of two discriminators are expressed by Eq.17 and 18.
\begin{equation}
{L_{{D_r}}} = \frac{1}{N}\sum\limits_{n = 1}^N {\left[ {{D_r}(I_{r,f}^n) + \lambda {{(1 - ||\nabla {D_r}(I_r^n)|{|_2})}^2}} \right]} \	
\end{equation}
\begin{equation}
{L_{{D_v}}} = \frac{1}{N}\sum\limits_{n = 1}^N {\left[ {{D_v}(I_{v,f}^n) + \lambda {{(1 - ||\nabla {D_v}(I_v^n)|{|_2})}^2}} \right]} \	
\end{equation}
where $ \lambda  $ is the regularization parameter, $ ||\cdot||_2 $ denotes  $ L_{2} $ norm. The first term represents the wasserstein distance between fused result and infrared or visible image, while the second term is the gradient penalty, which limits the learning ability of the discriminator.

\begin{figure*}[!t]
	\centering
	\includegraphics[width=1\textwidth]{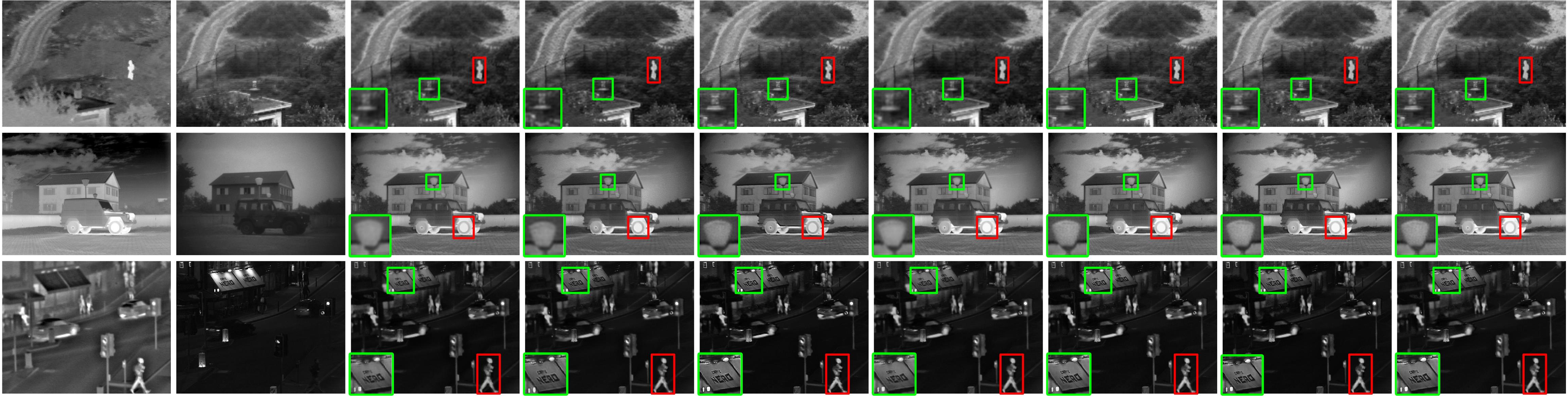}
	% where an .eps filename suffix will be assumed under latex, 
	% and a .pdf suffix will be assumed for pdflatex; or what has been declared
	% via \DeclareGraphicsExtensions.
	\caption{The subjective ablation results of attention mechanism for three typical examples. The first two columns are source images, and others are the fusion images obtained by No\_Attention, Only\_interact, Only\_VIS\_Com, Only\_IR\_Com, Only\_Channel, Only\_Spatial and our ICAFusion, respectively.}
	\label{Fig4}
\end{figure*}

\begin{table*}[!t]
	\renewcommand\arraystretch{1.5}
	%\renewcommand\tabcolsep{3.5pt}
	%% increase table row spacing, adjust to taste
	%\renewcommand{\arraystretch}{1.3}
	% if using array.sty, it might be a good idea to tweak the value of
	% \extrarowheight as needed to properly center the text within the cells
	\caption{The objective ablation experiments with different attention models on the TNO dataset.}
	\label{table4}
	\centering
	%% Some packages, such as MDW tools, offer better commands for making tables
	%% than the plain LaTeX2e tabular which is used here.
	\begin{tabular}{ l c c c c c c c c c }
		\hline
		Models  & AG & EN & SD & MI & SF & NCIE & Q$_{abf} $ & VIF     \\
		\hline
		No\_Attention & 3.16127 & 7.04056 & 39.10340 & 2.85232 & 6.34894 & 0.80681 & 0.31202 & 0.33340 	\\
		Only\_interact & 3.86125 & 7.02053 & 39.69077 & 2.72889 & 7.59874 & 0.80646 & 0.31936 & 0.34776       \\ 
		Only\_VIS\_Com & 5.66921 & 6.97240 & 37.70503 & 3.99192 & \underline{11.14457} & 0.81326 & 0.45124 & 0.44639   \\
		Only\_IR\_Com & 3.33456 & \underline{7.05532} & 39.70925 & 2.87210 & 6.75306 & 0.80687 & 0.34044 & 0.36369    \\
		Only\_Channel & \underline{5.80310} & 7.05136 & 39.87672 & \textbf{4.23417} & 11.10847 & \underline{0.81404} & \underline{0.47871} & \textbf{0.48691} 	 \\  
		Only\_Spatial & 5.69037 & 7.05013 & \underline{40.08709} & 4.17194 & 10.99383 & 0.81338 & 0.46603 & \underline{0.48567}		\\    
		Ours & \textbf{5.84108}  & \textbf{7.06216} & \textbf{40.26921} & \underline{4.23011} & \textbf{11.18681} & \textbf{0.81420} & \textbf{0.47935} & 0.48389 \\  
		\hline 		
	\end{tabular}
\end{table*}

\section{experiments and discussions}

In this section, the experimental settings are firstly described, and then the ablation study on attention mechanism is discussed. Finally, we conduct the related experiments on different datasets to demonstrate the effectiveness and superiority of our ICAFusion.

\subsection{Training and testing details}
In the training process, the TNO dataset [38] including 25 infrared and visible image pairs are proposed for the training. To expand the training dataset, we use the sliding step of 12 to divide original image pairs into the size of $ 128\times 128 $, and convert the gray value range to [-1, 1]. Thus, we can obtain 18813 patch pairs. In addition, The adam optimizer is applied to update model parameters, batchsize and epoch are set to 4 and 16, respectively. The learning rate of the generator and discriminator are set as $1 \times 10^{-4}$ and $4 \times  10^{-4}$, and the corresponding iterations are set to 1 and 2, respectively. In the loss function, the regularization parameter $ \lambda  $ is set to 10. The experimental training platform is Intel I9-10850K CPU, 64 GB memory and NVIDIA GeForce GTX 3090 GPU. The programming environment is Python and PyTorch platforms.

In the testing process, the TNO, Roadscene [39] and OTCBVS [40] datasets are used for the testing, in which 22, 28, 40 image pairs and Nato\_camp sequence are successively selected. We adopt nine representative methods, namely MDLatLRR [7], DenseFuse [15], IFCNN [17], Res2Fusion [25], SEDRFuse [23], RFN-Nest [28], PMGI [31], FusionGAN [19] and GANMcC [34], to compare with our ICAFusion. Besides, eight metrics, such as average gradient (AG), entropy (EN) [41], standard deviation (SD) [42], mutual information (MI) [43], spatial frequency (SF) [44], nonlinear correlation information entropy (NCIE) [44], Q$_{abf}$ [45] and visual information fidelity (VIF) [46] are employed for objective evaluation.

\subsection{Ablation study on attention mechanism}
In our fusion network, the interactive and compensatory attention modules are proposed to model the long-range dependencies from the channel and spatial dimensions, which are further used to interact and compensate features. To verify their effectiveness and superiority, we use six validation models for comparison, which are without attention modules, termed as No\_Attention, only retaining the interactive attention modules without compensatory attention modules, termed as Only\_interact, only retaining visible compensatory attention modules, termed as Only\_VIS\_Com, only retaining infrared compensatory attention modules, termed as Only\_IR\_Com, only retaining channel attention mechanism, termed as Only\_Channel and only retaining spatial attention mechanism, termed as Only\_Spatial. The optimal values are described in bold, while suboptimal values are underlined.

The subjective ablation results of three typical examples, such as \textit{Nato\_camp, Jeep} and \textit{Street}, are shown in Fig.4. By contrast, Only\_interact achieves better visual effect than that of No\_Attention. For example, for the \textit{Nato\_camp}, Only\_interact has higher brightness pedestrian and clear chimney details. This is because the interactive attention modules communicate their pathwise information of the triple path, and further improve feature representational capacity. Due to only a single modality compensatory information, Only\_VIS\_Com and Only\_IR\_Comp produce an unbalanced fusion result. Only\_VIS\_Com has clear texture details, and lost the brightness of infrared targets, while Only\_IR\_Com generates the opposite effect. Moreover, Only\_Channel and Only\_Spatial achieve similar results with our ICAFusion from the subjective visual observation. 

Table I presents the objective ablation experiments with different attention models on the TNO dataset. Compared with No\_Attention and Only\_interact, the former obtains the best metrics for EN, MI, NCIE, while the latter ahieves best metrics for AG, SD, SF, Q$_{abf}$ and VIF, indicating that our interactive attention modules are effective. In addition, Both Only\_VIS\_Com and Only\_IR\_Comp obtain better metrircs than No\_Attention, except that EN of Only\_VIS\_Com is lower than that of No\_Attention. This explains that the compensatory attention modules can compensate infrared pixel intensity and visible texture details for feature reconstruction. Only\_Channel and Only\_Spatial yields average values of metrics close to our method. However, our ICAFusion acquires the first rank for AG, EN, SD, SF, NCIE and Q$_{abf}$, the second and third ranks for MI and VIF, indicating that the proposed method has better fusion performance, and the proposed attention mechanism is effective and reasonable.

\begin{figure*}[!t]
	\centering
	\includegraphics[width=0.94\textwidth]{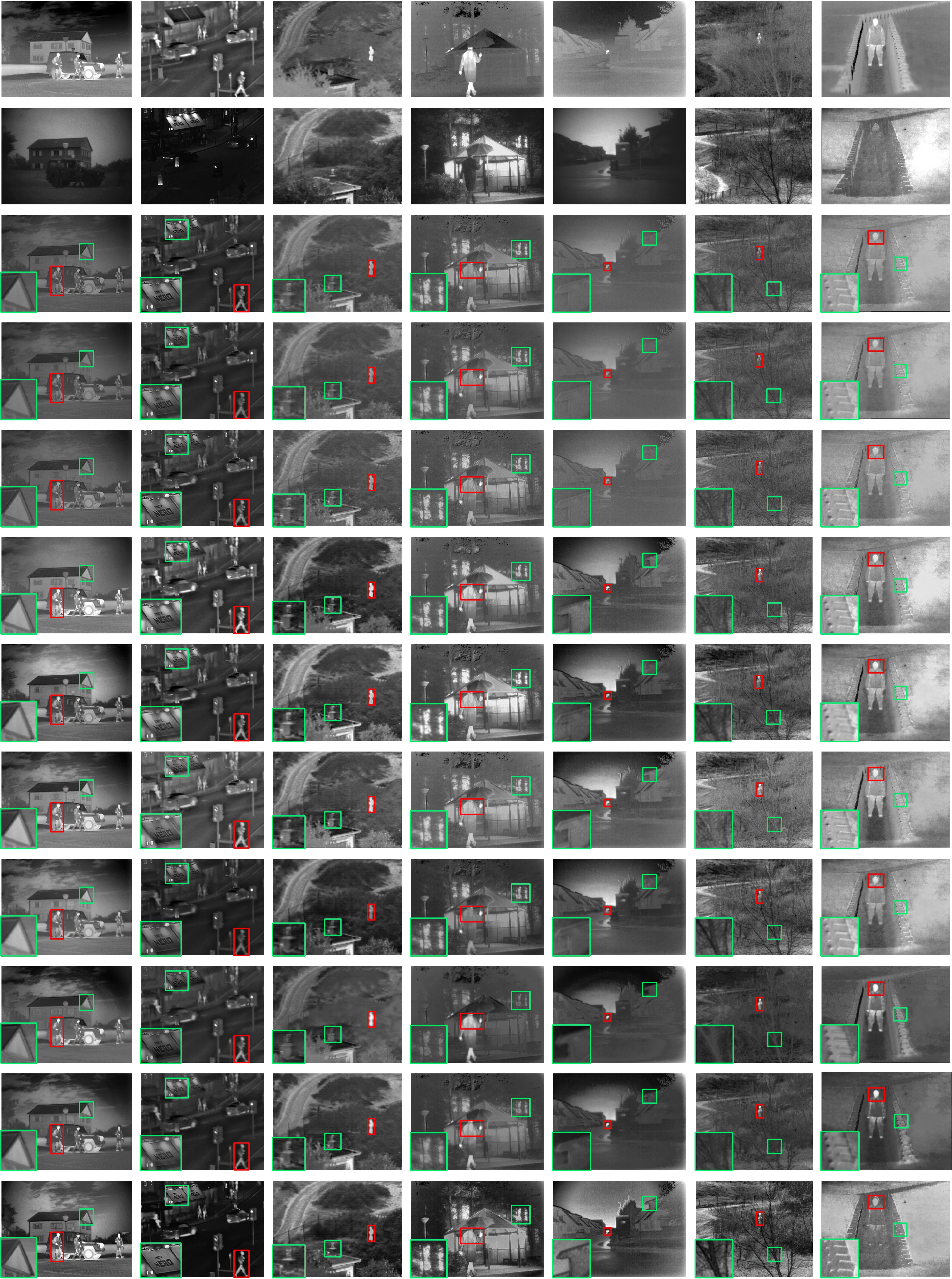}
	% where an .eps filename suffix will be assumed under latex, 
	% and a .pdf suffix will be assumed for pdflatex; or what has been declared
	% via \DeclareGraphicsExtensions.
	\caption{The subjecive comparative results of seven typical examples selected from TNO dataset, such as \textit{Soldiers\_with\_jeep, Street, Nato\_camp, Kaptein\_1654, Movie\_01, Sandpath} and \textit{soldier\_in\_trench\_1}. The top two lines are source images, and others are the fusion images obtained by MDLatLRR [7], DenseFuse [15], IFCNN [17], Res2Fusion [25], SEDRFuse [23], RFN-Nest [28], PMGI [31], FusionGAN [19], GANMcC [34] and our ICAFusion, respectively.}
	\label{Fig5}
\end{figure*}

\begin{figure*}[!t]
	\centering
	\includegraphics[width=0.995\textwidth]{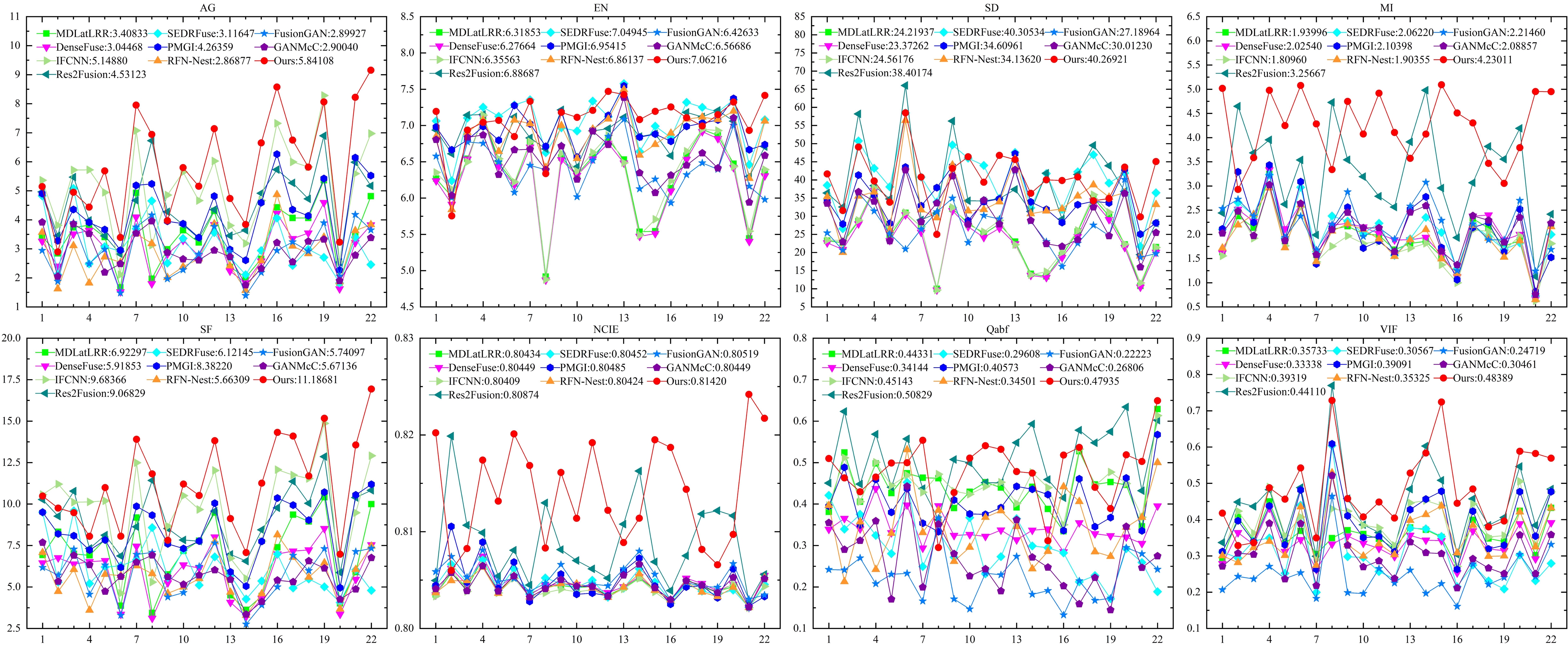}
	% where an .eps filename suffix will be assumed under latex, 
	% and a .pdf suffix will be assumed for pdflatex; or what has been declared
	% via \DeclareGraphicsExtensions.
	\caption{The subjective comparative results of eight evalution metrics for TNO dataset. The corresponding average values of different fusion methods are also presented. Note that our ICAFusion is indicated by a red dotted line.}
	\label{Fig6}
\end{figure*}

\begin{figure*}[!t]
	\centering
	\includegraphics[width=0.995\textwidth]{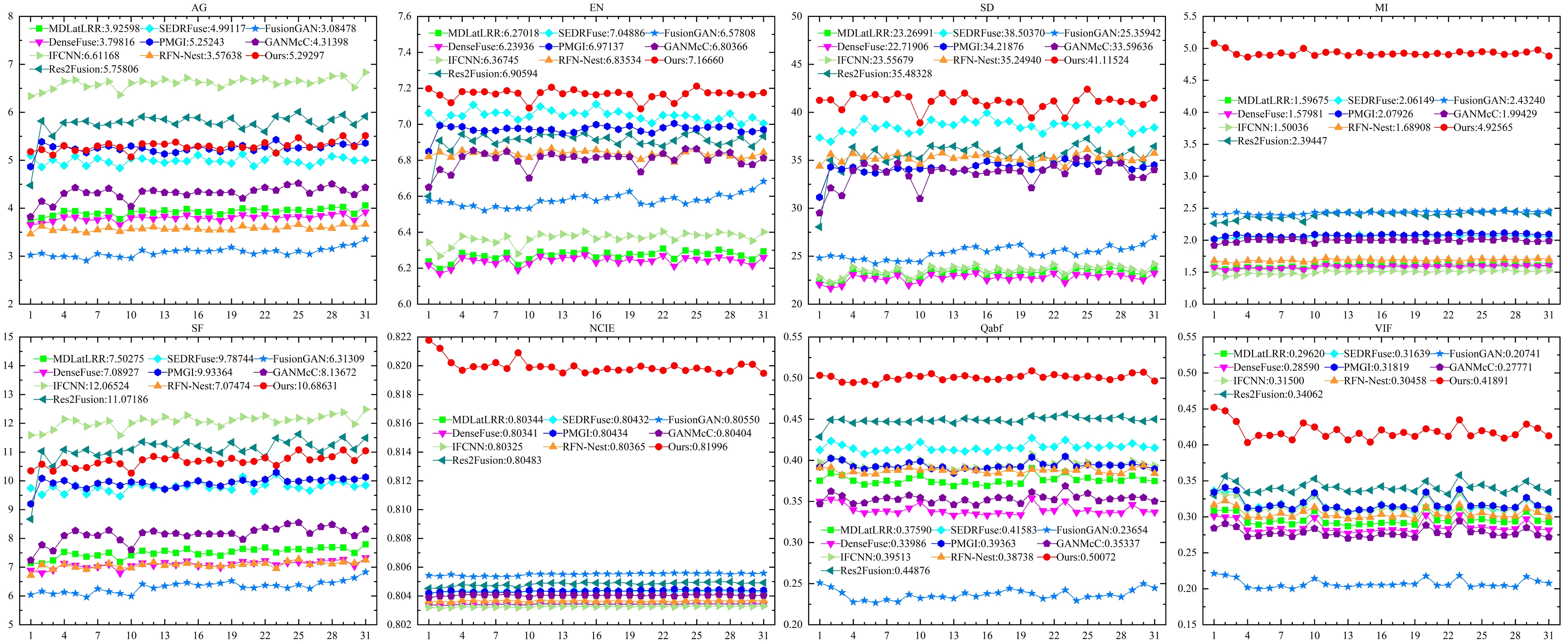}
	% where an .eps filename suffix will be assumed under latex, 
	% and a .pdf suffix will be assumed for pdflatex; or what has been declared
	% via \DeclareGraphicsExtensions.
	\caption{The subjective comparative results of eight evalution metrics for Nato\_camp sequence. The corresponding average values of different fusion methods are also presented. Note that our ICAFusion is indicated by a red dotted line.}
	\label{Fig7}
\end{figure*}

\subsection{Results on TNO dataset}
We conduct the experiments on TNO dataset to demonstrate the effectiveness of the proposed ICAFusion. Seven typical image pairs, such as \textit{Soldiers\_with\_jeep, Street, Nato\_camp, Kaptein\_1654, Movie\_01, Sandpath} and \textit{soldier\_in\_trench\_1}, are choosed for the subjective validation, and the corresponding comparative results are presented in Fig.5. From these results, the traditional method MDLatLRR proposes the learnbale low-rank respresentation, the obtained fused results exist undesired artifacts. The CNN-based methods, such as DenseFuse and IFCNN, apply average fusion rule under the simple network framework, the obtaied results have obvious detail missing and low contrast. However, SEDRFuse and Res2Fusion achieve relatively better performance because these methods propose fusion strategy based on attention mechanism. Their results can retain typical infrared targets, but produce some sharpened effects in certain degree, and some useful texture information is lost. In addition, for the end-to-end methods, RFN-Nest is inclined to preserve abundant visible details while missing typical infrared targets. PMGI achieves satisfactory results by maintaining the proportional of gradient and intensity, but its ability to perceive infrared targets and characterize visible details is still limited. FusionGAN and GANMcC intend to retain more prominent target information from infrared images. Due to a discriminator, FusionGAN achieves unbalanced results, which sharpens the infrared target edges and lacks the important visible details. Although GANMcC proposes two discriminators to realize some visual improvement, some useful texure details of visible images are still missing. Compared with the above methods, our ICAFusion achieves the optimal visual effects in simultaneously maintaining typical infrared targets and unambiguous visible details.

To facilitate visual observation, we mark some typical infrared targets in the red box, and magnify the representative visible details in the green box. As shown in Fig.5, for the first column images, $ i.e.,$ the results of \textit{Soldiers\_with\_jeep}, MDLatLRR, DenseFuse, IFCNN and RFN-Nest can preserve the texture details of the housetop, but lost the brightness of pedestrian. On the contrary, FusionGAN and GANMcC can retain the targets of infrared images, while the edges of pedestrians  are blurred, and the details of the housetop are missing. SEDRFuse and Res2Fusion achieve better results, but their visual effects are also limited. Specially, Res2Fusion lacks some useful scene information, such as trees and cloud. For the results of \textit{Street}, compared with other methods, our ICAFusion can preserve higher brightness of pedestrian and clearer details of billboard, and our result has higher image contrast. The other five image pairs can draw a similar conclusion. In general, the objective experiments demonstrate that our method can obtain better image fusion performance, and the generated results are more appropriate to the human visual system.

\begin{figure*}[!t]
	\centering
	\includegraphics[width=1\textwidth]{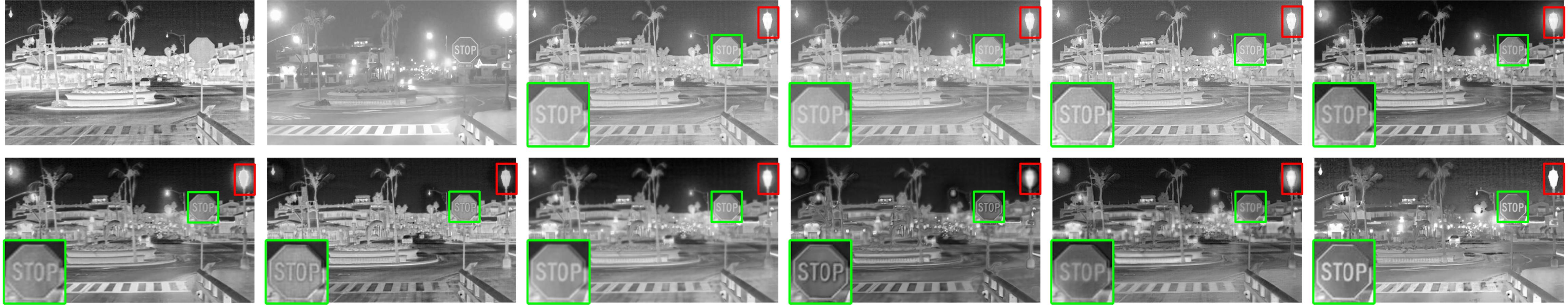}
	% where an .eps filename suffix will be assumed under latex, 
	% and a .pdf suffix will be assumed for pdflatex; or what has been declared
	% via \DeclareGraphicsExtensions.
	\caption{The subjecive comparative results of \textit{FLIR\_07210} selected from Roadscene dataset for different fusion methods. The left two images are source images, and others are the fusion images obtained by MDLatLRR [7], DenseFuse [15], IFCNN [17], Res2Fusion [25], SEDRFuse [23], RFN-Nest [28], PMGI [31], FusionGAN [19], GANMcC [34] and our ICAFusion, respectively.}
	\label{Fig8}
\end{figure*}

\begin{figure*}[!t]
	\centering
	\includegraphics[width=1\textwidth]{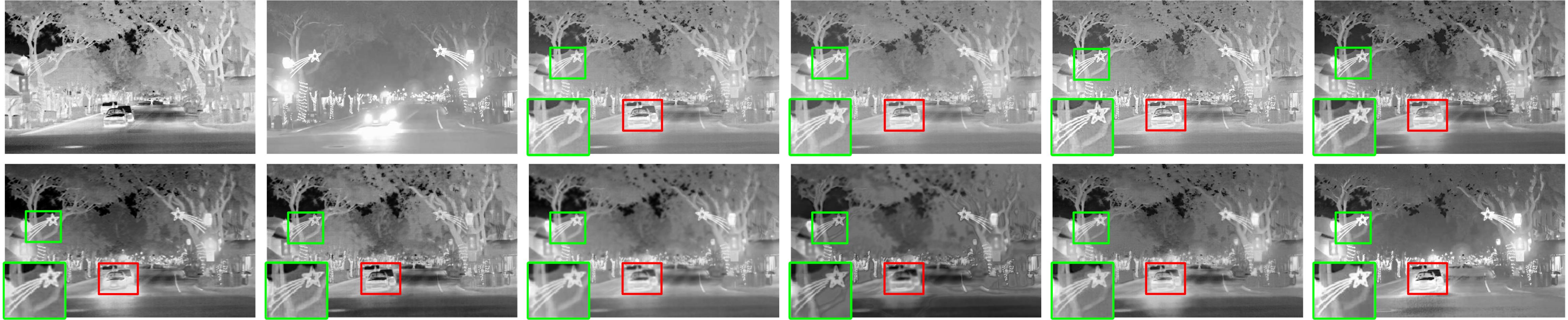}
	% where an .eps filename suffix will be assumed under latex, 
	% and a .pdf suffix will be assumed for pdflatex; or what has been declared
	% via \DeclareGraphicsExtensions.
	\caption{The subjecive comparative results of \textit{FLIR\_07081} selected from Roadscene dataset for different fusion methods. The left two images are source images, and others are the fusion images obtained by MDLatLRR [7], DenseFuse [15], IFCNN [17], Res2Fusion [25], SEDRFuse [23], RFN-Nest [28], PMGI [31], FusionGAN [19], GANMcC [34] and our ICAFusion, respectively.}
	\label{Fig9}
\end{figure*}

\begin{figure*}[!t]
	\centering
	\includegraphics[width=1\textwidth]{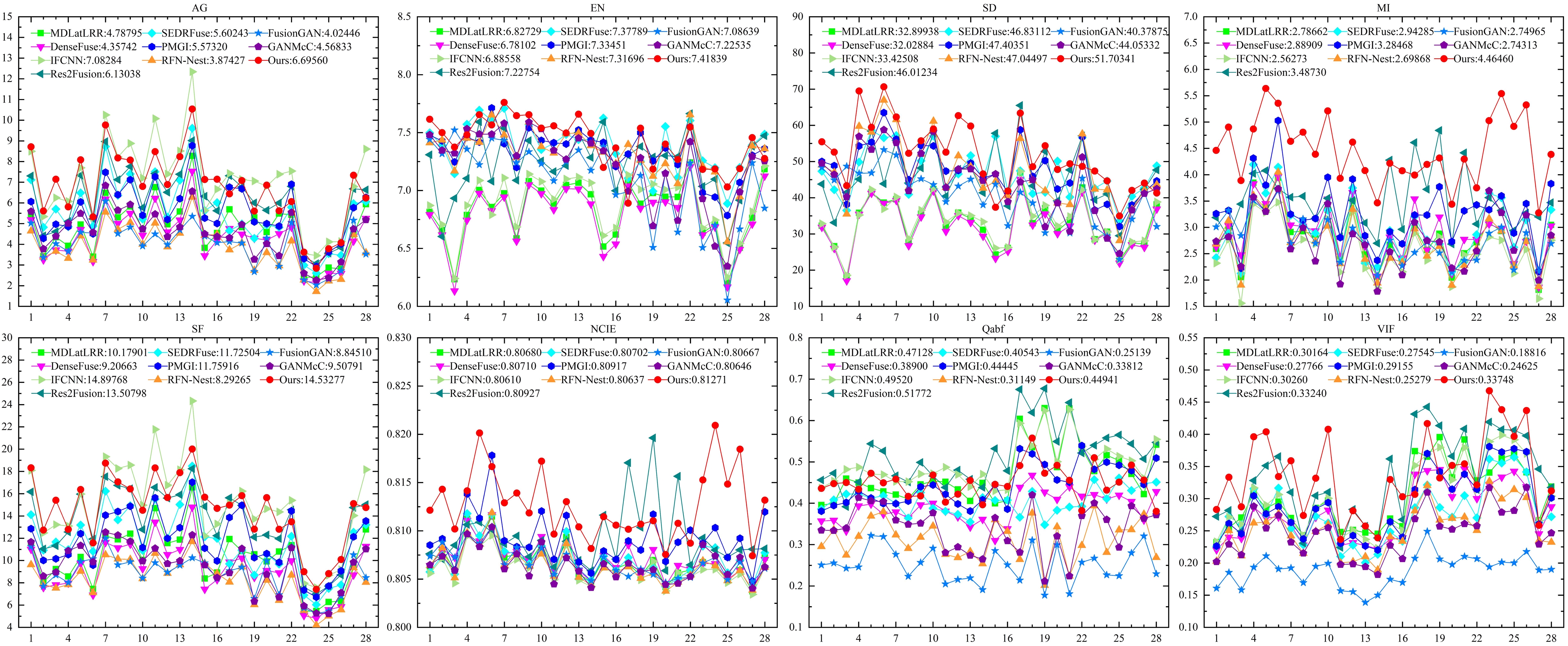}
	% where an .eps filename suffix will be assumed under latex, 
	% and a .pdf suffix will be assumed for pdflatex; or what has been declared
	% via \DeclareGraphicsExtensions.
	\caption{The subjective comparative results of eight evalution metrics for Roadscene dataset. The corresponding average values of different fusion methods are also presented. Note that our ICAFusion is indicated by a red dotted line.}
	\label{Fig10}
\end{figure*}

We continue to verify our ICAFusion from the perspective of objective evaluation. Figure 6 gives the comparative results of different methods for TNO dataset. Note that our metric curves are described by a red dotted line, and the average values of each metric for different methods are also presented. We can find that our ICAFusion achieves the highest values of most metrics for each image pair. Meanwhile, our ICAFusion acquires the first rank for AG, EN, MI, SF, NCIE and VIF, and the second rank for SD and Q$_{abf}$, which follow behind IFCNN and Res2Fusion, respectively. In addition, the subjective comparative results of the Nato\_camp sequence are shown in Fig.7. Our ICAFusion acquires the first rank for EN, SD, MI, NCIE, Q$_{abf}$ and VIF, and the third rank for AG and SF, which are lower than IFCNN and Res2Fusion. In conclusion, our ICAFusion implements higher performance, and surpasses other representative methods in the subjective visual description and objective metric evaluation.

\begin{figure*}[!t]
	\centering
	\includegraphics[width=1\textwidth]{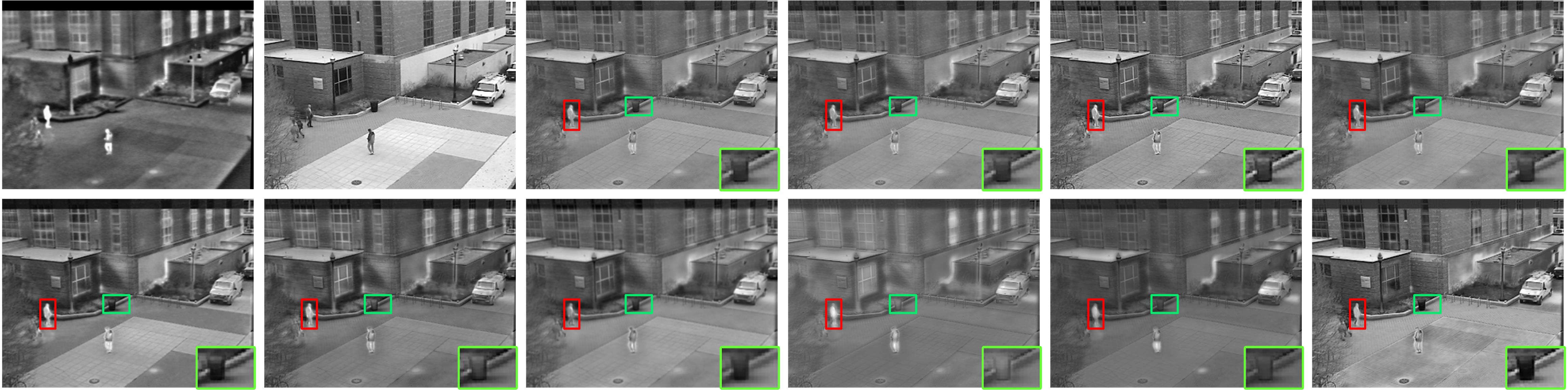}
	% where an .eps filename suffix will be assumed under latex, 
	% and a .pdf suffix will be assumed for pdflatex; or what has been declared
	% via \DeclareGraphicsExtensions.
	\caption{The subjecive comparative results of \textit{video\_1007} selected from OTCBVS dataset for different fusion methods. The left two images are source images, and others are the fusion images obtained by MDLatLRR [7], DenseFuse [15], IFCNN [17], Res2Fusion [25], SEDRFuse [23], RFN-Nest [28], PMGI [31], FusionGAN [19], GANMcC [34] and our ICAFusion, respectively.}
	\label{Fig11}
\end{figure*}
\begin{figure*}[!t]
	\centering
	\includegraphics[width=1\textwidth]{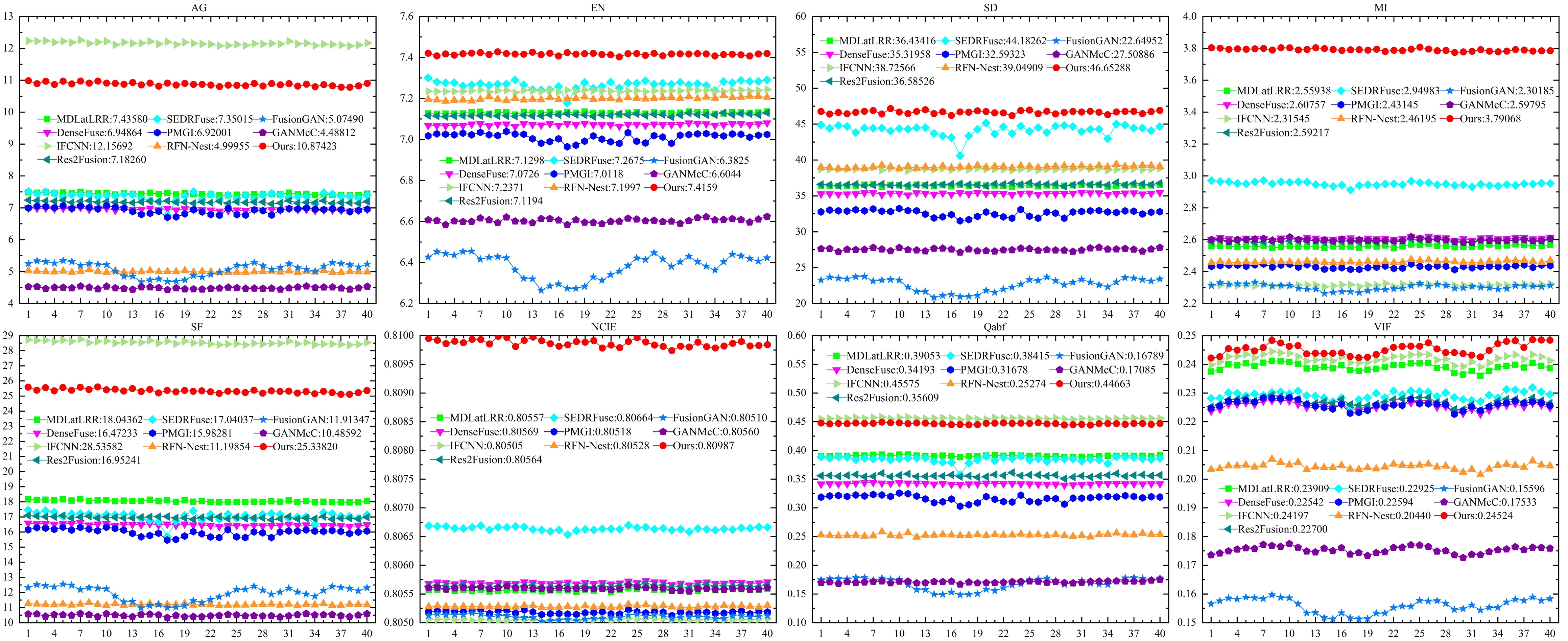}
	% where an .eps filename suffix will be assumed under latex, 
	% and a .pdf suffix will be assumed for pdflatex; or what has been declared
	% via \DeclareGraphicsExtensions.
	\caption{The subjective comparative results of eight evalution metrics for OTCBVS dataset. The corresponding average values of different fusion methods are also presented. Note that our ICAFusion is indicated by a red dotted line.}
	\label{Fig12}
\end{figure*}

\begin{table}[!t]
	\renewcommand\arraystretch{1.5}
	\scriptsize
	%% increase table row spacing, adjust to taste
	%\renewcommand{\arraystretch}{1.3}
	% if using array.sty, it might be a good idea to tweak the value of
	% \extrarowheight as needed to properly center the text within the cells
	\caption{The comparative results of fusion computational efficiency for three datasets (Unit: second).}
	\label{table8}
	\centering
	%% Some packages, such as MDW tools, offer better commands for making tables
	%% than the plain LaTeX2e tabular which is used here.
	\begin{tabular}{ l l l l}
		\hline
		Method & TNO & Roadscene & OTCBVS \\
		\hline
		MDLatLRR  & 7.941$\times10^{1} $ & 2.441$\times 10^{1}$ & 3.839$\times 10^{1}$  \\
		DenseFuse  & 8.509$\times 10^{-2}$ & 2.893$\times 10^{-2}$ & 4.001$\times 10^{-2}$  \\	
		SEDRFuse  & 2.676 & 1.445 & 8.031$\times 10^{-1}$ \\
		Res2Fusion  & 1.886$\times 10^{1}$ & 4.267 & 1.337  \\
		IFCNN  & 4.554$\times 10^{-2}$ & 2.246$\times 10^{-2}$ & 1.149$\times 10^{-2}$   \\
		PMGI  & 5.445$\times 10^{-1}$ & 2.928$\times 10^{-1}$ & 1.262$\times 10^{-1}$  \\
		RFN-Nest & 1.777$\times 10^{-1}$  & 8.609$\times 10^{-2}$ & 5.181$\times 10^{-2}$   \\
		FusionGan& 2.015 & 1.093 &  4.903$\times 10^{-1}$	 \\
		GanMcC & 4.210 &  2.195 & 1.017   \\
		Ours  & 1.309$\times 10^{-1}$ & 7.610$\times 10^{-2}$ & 3.245$\times 10^{-2}$  \\
		\hline 
	\end{tabular}
\end{table}

\subsection{Results on Roadscene dataset}
To further illustrate the superiority of the proposed method, 28 infrared and visible image pairs are selected from the Roadscene dataset for experimental verification. Fig.8 and 9 give the subjective comparative results with different methods for \textit{FLIR\_07210} and \textit{FLIR\_07081}. These results indicate that our ICAFusion owns three distinct advantages. Firstly, our method can retain the high-brigtness target information from the infrared image. As shown in Fig.7 and 8, for typical infrared targets, $ e.g.,$ street lamp and car, our results have higher brightness than other methods. Secondly, our method can perserve abundant and unambiguous texuture details from the visible image. For example, the representational details, $ e.g.,$ signboard and decorative lights, obtained by our method are more obvious and clearer than that of other methods. Thirdly, our method can achieve higher contrast and better visual perception. Compared with source images and other fused results, due to the application of interactive and compensatory attention modules, the proposed ICAFusion can well preserve prominent target characteristics and unambiguous scene details in the fusion images.

Meanwhile, Fig.10 shows the objective results of different methods for the Roadscene dataset. the proposed method obtains the first rank for metrics EN, SD, MI, NCIE and VIF, the second rank for metrics AG, SF, which are only in arrears of IFCNN. The objective experiments also demonstrate that the fusion performance of our ICAFusion surpasses other methods. In addition, the largest value EN indicates that our results can maintain abdundant useful information from source images. This is because our method proposes a triple path where infrared and visible paths can provide additional intensity and gradient information for the fused image. The largest MI and NCIE demonstrate that our results have a strong correlation and similarity with source images. The reason is that our method adopts two discriminators to supervise and optimize the generator with a specific loss function, and can produce a more balanced fusion result. The largest SD and VIF explain that our results can achieve better image contrast and visual effect. This is because our interactive and compensatory attention modules can model the long-range dependencies, and refine features to more place emphasis on infrared target perception and visible detail characterization.

\subsection{Results on OTCBVS dataset}
We further carry on the experiments on the OTCBVS dataset to clarify the generalization ability of our ICAFusion. We select 40 image pairs of the pedestrian change sequence, and the comparative results are shown in Fig.11. By contrast, our ICAFusion presents a more richer background scene, and involves unambiguous details of the ash-bin. The typical target region, e.g., the pedestrians, can also be contained. As a whole, our method generates a more balanced result and produces better visual perception. The corresponding objective comparative results are shown in Fig.12. Our method acquires the first rank for EN, SD, MI, NCIE and VIF, and the second rank for AG, SF and Qabf , which only follows behind IFCNN.

In order to verify the fusion computational eﬀiciency, the traditional method MDLatLRR is tested on the CPU, while the others are implemented on the GPU. Table II shows the comparative results of different fusion methods. The experiments show that our ICAFusion achieves the competitive fusion eﬀiciency, which is slightly lower than that of DenseFuse and IFCNN. The main reason is that both methods propose a simple network framework with a weighted average fusion rule. In conclusion, the above subjective and objective experiments demonstrate that our ICAFusion achieves remarkable results, and is superior to other methods on different datasets, indicating that it has better fusion performance and stronger generalization ability.

\section{Conclusion}
In this paper, an interactive compensatory attention adversarial learning network, termed as ICAFusion, is developed. We construct a multi-level encoder-decoder network with a triple path, and infrared and visible paths provide additional intensity and gradient information for the subsequent processing. The interactive and compensatory attention modules are developed to communicates their pathwise information and model the long-range dependencies. The obtained attention maps can more emphasis on infrared target perception and visible detailed characterization, and further increase the representation power of feature extraction and feature reconstruction. In addition, dual discriminators are designed to identify the similar distribution between fused result and source images. Moreover, the specific loss function is adopted, and optimize the generator to produce a more balanced result.

We carry out extensive experiments on the TNO, Roadscene and OTCBVS datasets, and the related results demonstrate that our ICAFusion achieves satisfactory fusion performance along with high computational eﬀiciency and strong generalization ability, preceding other nine state-of-the-art fusion methods in the subjective visual description and objective metric evaluation. In the future work, we will continue to optimize the network architecture, and introduce attention mechanisms into discriminator to further improve the equilibrium and effectiveness of the adversarial training. Meanwhile, we will also extend this network for other tasks, such as multi-band, multi-exposure and multi-focus image fusion, etc.

% or


%\appendices

%\section{}

%\section*{Acknowledgment}
%This work is supported by the Applied Basic Research Project of Shanxi Province under Grant 201901D111260, the Open Foundation of Shanxi Key Laboratory of Signal Capturing \& Processing under Grant ISPT2020-4, the Startup Foundation for Doctors of Taiyuan University of Science and Technology under Grant 20162004.

\ifCLASSOPTIONcaptionsoff
  \newpage
\fi



%\bibliographystyle{IEEEtran}
% argument is your BibTeX string definitions and bibliography dataset(s)

\begin{thebibliography}{1}

\bibitem{9340007}
Q.~Xu, Y.~Mei, J.~Liu and C.~Li, ``Multimodal cross-layer bilinear pooling for RGBT tracking,'' \emph{IEEE Trans. Multimedia}, 2021. doi: 10.1109/TMM.2021.3055362.
	
\bibitem{9454273}
W.~Zhou, Y.~Zhu, J.~Lei, J.~Wan and L.~Yu, ``CCAFNet: Crossflow and cross-scale adaptive fusion network for detecting salient objects in RGB-D images,'' \emph{IEEE Trans. Multimedia}, 2021. doi: 10.1109/TMM.2021.3077767.

\bibitem{8765608}
Z.~Feng, J.~Lai and X.~Xie, ``Learning modality-specific representations for visible-infrared person re-identification,'' \emph{IEEE Trans. Image Process.}, vol.~29, pp.~579-590, 2020.

\bibitem{2015Multi}
Z.~Wang, F.~Yang, Z.~Peng, L.~Chen, and L.~Ji, ``Multi-sensor image enhanced fusion algorithm based on NSST and top-hat transformation,'' \emph{Optik}, vol. 126, no.~23, pp. 4184-4190, 2015.

\bibitem{0Infrared}
Z.~Wang, J.~Xu, X.~Jiang, and X.~Yan, ``Infrared and visible image fusion via hybrid decomposition of NSCT and morphological sequential toggle operator,'' \emph{Optik}, vol. 201, no.~1, p. 163497, 2020.

\bibitem{2018Sparse}
Q.~Zhang, Y.~Liu, R.~S. Blum, J.~Han, and D.~Tao, ``Sparse representation based multi-sensor image fusion for multi-focus and multi-modality images: A review,'' \emph{Inf. Fusion}, vol.~40, pp. 57-75, 2018.

\bibitem{9018389}
H.~Li, X.~Wu and J.~Kittler, ``MDLatLRR: A novel decomposition method for infrared and visible image fusion,'' \emph{IEEE Trans. Image Process.}, vol.~29, pp.~4733-4746, 2020.

\bibitem{2017Infrared}
J.~Ma, Z.~Zhou, B.~Wang, and H.~Zong, ``Infrared and visible image fusion based on visual saliency map and weighted least square optimization,''
\emph{Infr. Phys. Technol.}, vol.~82, pp. 8-17, 2017.

\bibitem{KONG2014161}
W.~Kong, L.~Yang, X.~Feng and H.~Zhao, ``Adaptive fusion method of visible light and infrared images based on non-subsampled shearlet transform and fast non-negative matrix factorization,'' \emph{Infr. Phys. Technol.}, vol.~67, pp.~161-172, 2014.

\bibitem{HU2019102977}
P.~Hu, F.~Yang, H.~Wei, L.~Ji and D.~Liu, ``A multi-algorithm block fusion method based on set-valued mapping for dual-modal infrared images,'' \emph{Infr. Phys. Technol.}, vol.~102, pp.~102977, 2019.

\bibitem{9340007}
Z.~Li, H.~Hu, W.~Zhang, S.~Pu and B.~Li, ``Spectrum characteristics preserved visible and near-infrared image fusion algorithm,'' \emph{IEEE Trans. Multimedia}, vol.~23, pp.~306-319, 2021.

\bibitem{9376921}
R.~Nie, C.~Ma, J.~Cao, H.~Ding and D.~Zhou, ``A total variation with joint norms for infrared and visible image fusion,'' \emph{IEEE Trans. Multimedia}, 2021. doi: 10.1109/TMM.2021.3065496.

\bibitem{ZHANG2021323}
H.~Zhang, H.~Xu, X.~Tian, J.~Jiang and J.~Ma, ``Image fusion meets deep learning: A survey and perspective,'' \emph{Inf. Fusion}, vol.~76, pp. 323-336, 2021.

\bibitem{9479765}
L.~Jian, R.~Rayhana, L.~Ma, S.~Wu, Z.~Liu and H.~Jiang, ``Infrared and visible image fusion based on deep decomposition network and saliency analysis,'' \emph{IEEE Trans. Multimedia}, 2021. doi: 10.1109/TMM.2021.3096088.

\bibitem{8580578}
H.~Li and X.~Wu, ``Densefuse: A fusion approach to infrared and visible
images,'' \emph{IEEE Trans. Image Process.}, vol.~28, no.~5, pp.
2614-2623, 2019.

\bibitem{9622164}
X.~Luo, Y.~Gao, A.~Wang, Z.~Zhang and X.~Wu, ``IFSepR: A general framework for image fusion based on separate representation learning,'' \emph{IEEE Trans. Multimedia}, 2021. doi: 10.1109/TMM.2021.3129354.

\bibitem{2020IFCNN}
Y.~Zhang, Y.~Liu, P.~Sun, H.~Yan, X.~Zhao, and L.~Zhang, ``Ifcnn: A general
image fusion framework based on convolutional neural network,''
\emph{Inf. Fusion}, vol.~54, pp. 99-118, 2020.

\bibitem{10.1007/978-3-319-10602-1_48}
T.~Lin, M.~Maire, S.~Belongie, J.~Hays, P.~Perona, D.~Ramanan,
P.~Doll{\'a}r and C.~L. Zitnick, ``Microsoft coco: Common objects in
context,'' in \emph{Computer Vision -- ECCV 2014}, D.~Fleet, T.~Pajdla, B.~Schiele, and T.~Tuytelaars, Eds.\hskip 1em plus 0.5em minus 0.4em\relax Cham: Springer International Publishing, 2014, pp. 740-755.

\bibitem{2019FusionGAN}
J.~Ma, W.~Yu, P.~Liang, C.~Li and J.~Jiang, ``Fusiongan: A generative
adversarial network for infrared and visible image fusion,''
\emph{Inf. Fusion}, vol.~48, pp.~11-26, 2019.

\bibitem{2020Infrared}
J.~Ma, P.~Liang, W.~Yu, C.~Chen, X.~Guo, J.~Wu, and J.~Jiang, ``Infrared and visible image fusion via detail preserving adversarial learning,''
\emph{Inf. Fusion}, vol.~54, pp. 85-98, 2020.

\bibitem{9274337}
Y.~Fu, X.~Wu and T.~Durrani, ``Image fusion based on generative adversarial network consistent with perception,'' \emph{Inf. Fusion}, vol.~72, pp.~110-125, 2021.

\bibitem{3305404}
M.~Arjovsky, S.~Chintala, L.~Bottou, ``Wasserstein generative adversarial networks,'' in \emph{Proc. Inter. Conf. Mach. Learn. (ICML)}, Sydney, Australia, Aug.~2017, vol.~70, pp.~214–223.

\bibitem{2020SEDRFuse}
L.~Jiang, X.~Yang, Z.~Liu, G.~Jeon, M.~Gao and D.~Chisholm, ``SEDRFuse: A symmetric encoder-decoder with residual block network for infrared and visible image fusion,'' \emph{IEEE Trans. Instrum. Meas.}, vol.~70, pp.~1-15, 2021.

\bibitem{2020NestFuse}
H.~Li, X.~Wu and T.~Durrani, ``Nestfuse: An infrared and visible image
fusion architecture based on nest connection and spatial/channel attention models,'' \emph{IEEE Trans. Instrum. Meas.}, vol.~69, no.~12, pp.~9645-9656, 2020.

\bibitem{9670874}
Z.~Wang, Y.~Wu, J.~Wang,J.~Xu and W.~Shao, ``Res2Fusion: Infrared and visible image fusion based on dense Res2net and double non-local
attention models,'' \emph{IEEE Trans. Instrum. Meas.}, 2021. doi: 10.1109/TIM.2021.3139654

\bibitem{9528393}
Z.~Wang, J.~Wang, Y.~Wu, J.~Xu and X.~Zhang, ``UNFusion: A unified multi-scale densely connected network for infrared and visible image fusion,'' \emph{IEEE Trans. Circuits Syst. Video Technol.}, 2021. doi: 10.1109/TCSVT.2021.3109895.

\bibitem{LONG2021128}
Y.~Long, H.~Jia, Y.~Zhong, Y.~Jiang and Y.~Jia, ``RXDNFuse: A aggregated residual dense network for infrared and visible image fusion,''
\emph{Inf. Fusion}, vol.~69, pp.~128-141, 2021.

\bibitem{LI202172}
H.~Li, X.~Wu and J.~Kittler, ``RFN-Nest: An end-to-end residual fusion network for infrared and visible images,'' \emph{Inf. Fusion}, vol.~73, pp.~72-86, 2021.

\bibitem{9167484}
F.~Zhao and W.~Zhao, ``Learning specific and general realm feature representations for image fusion,'' \emph{IEEE Trans. Multimedia}, vol.~23, pp.~2745-2756, 2021.


\bibitem{2020U2Fusion}
H.~Xu, J.~Ma, J.~Jiang, X.~Guo and H.~Ling, ``U2fusion: A unified unsupervised image fusion network,'' \emph{IEEE Trans. Pattern Anal. Mach. Intell.}, 2020, doi:10.1109/TPAMI.2020.3012548.

\bibitem{2020Rethinking}
H.~Zhang, H.~Xu, Y.~Xiao, X.~Guo and J.~Ma, ``Rethinking the image fusion: A fast unified image fusion network based on proportional maintenance of gradient and intensity,'' in \emph{Proc. AAAI Conf. Artif. Intell.}, vol.~34, no.~7, pp.~12797-12804, 2020.


\bibitem{9031751}
J.~Ma, H.~Xu, J.~Jiang, X.~Mei and X.~Zhang, ``DDcGAN: A dual-discriminator conditional generative adversarial network for multi-resolution image fusion,''
\emph{IEEE Trans. Image Process.}, vol.~29, pp.~4980-4995, 2020.

\bibitem{9623476}
H.~Zhou, W.~Wu, Y.~Zhang, J.~Ma and H.~Ling, ``Semantic-supervised infrared and visible image fusion via a dual-discriminator generative adversarial network,'' \emph{IEEE Trans. Multimedia}, 2021. doi: 10.1109/TMM.2021.3129609.

\bibitem{9274337}
J.~Ma, H.~Zhang, Z.~Shao, P.~Liang and H.~Xu, ``GANMcC: A generative adversarial network with multiclassification constraints for infrared and visible image fusion,'' \emph{IEEE Trans. Instrum. Meas.}, vol.~70, pp.~1-14, 2021.

\bibitem{9216075}
J.~Li, H.~Huo, C.~Li, R.~Wang, C.~Sui and Z.~Liu, ``Multigrained attention network for infrared and visible image fusion,'' \emph{IEEE Trans. Instrum. Meas.}, vol.~70, pp.~1-12, 2021.

\bibitem{9670874}
J.~Li, H.~Huo, C.~Li, R.~Wang and Q.~Feng, ``AttentionFGAN: Infrared and visible image fusion using attention-based generative adversarial networks,'' \emph{IEEE Trans. Multimedia}, vol.~23, pp.~1383-1396, 2021.

\bibitem{9670854}
S.~Woo, J.~Park, J.-Y.~Lee, and I.~S.~Kweon, ``CBAM: Convolutional block
attention module,'' in \emph{Proc. Eur. Conf. Comput. Vis. (ECCV)}, Munich,
Germany, Sep.~2018, pp.~3–19.


\bibitem{2014TNO}
A.~Toet(2014).\emph{TNO Image Fusion Dataset}. Figshare.Data.[Online]. 
Available: https://figshare.com/articles/TN\_Image\_Fusion\_Dataset/1008029.

\bibitem{2020Roadscene}
H.~Xu(2020). \emph{Roadscene Database}. [Online]. Available: https://
github.com/hanna-xu/RoadScene.

\bibitem{2020Roadscene}
S.~Ariffin(2016). \emph{OTCBVS Database}. [Online]. Available: http://vcipl-okstate.org/pbvs/bench/.

\bibitem{2008Assessment}
V.~Aslantas and E.~Bendes, ``Assessment of image fusion procedures using entropy, image quality, and multispectral classification,'' \emph{J. Appl. Remote Sens.}, vol.~2, no.~1, pp. 1-28, 2008.

\bibitem{1997In}
Y.~Rao, ``In-fibre bragg grating sensors,'' \emph{Meas. Sci. Technol.}, vol.~8, no.~4, pp. 355-375, 1997.

\bibitem{477498}
A.~Eskicioglu and P.~Fisher, ``Image quality measures and their performance,'' \emph{IEEE Trans. Commun.}, vol.~43, no.~12, pp.~2959-2965, 1995.

\bibitem{2011Objective}
Z.~Liu, E.~Blasch, Z.~Xue, J.~Zhao, R.~Laganiere and W.~Wu, ``Objective assessment of multiresolution image fusion algorithms for context enhancement in night vision: A comparative study,'' \emph{IEEE Trans. Pattern	Anal. Mach. Intell.,} vol.~34, no.~1, pp. 94-109, 2011.

\bibitem{Piella2003ANQ}
G.~Piella and H.~Heijmansu, ``A new quality metric for image fusion,'' in \emph{Proc. IEEE Int. Conf. Image Process., (ICIP)}, 2003, pp.~173-176.

\bibitem{2013A}
Y.~Han, Y.~Cai, Y.~Cao, and X.~Xu, ``A new image fusion performance metric
based on visual information fidelity,'' \emph{Inf. Fusion}, vol.~14,
no.~2, pp. 127-135, 2013.


	



\end{thebibliography}


\begin{IEEEbiography}[{\includegraphics[width=1in,height=1.25in,clip,keepaspectratio]{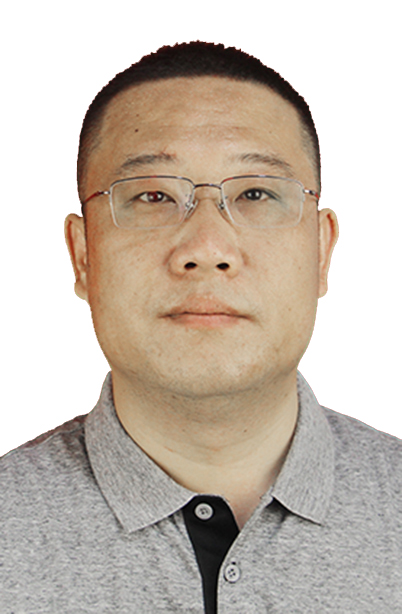}}]{Zhi-She Wang} \textit{(Member,~IEEE)} received the B.S degree in automation from North China Institute of Technology, Taiyuan, China, in 2002. He received the M.S. and Ph.D. degree in signal and information processing from the North University of China, Taiyuan, China, in 2007 and 2015. He is currently an associate professor with Taiyuan University of Science and Technology. His current research interests include computer vision, pattern recognition and machine learning.
\end{IEEEbiography}

\begin{IEEEbiography}[{\includegraphics[width=1in,height=1.25in,clip,keepaspectratio]{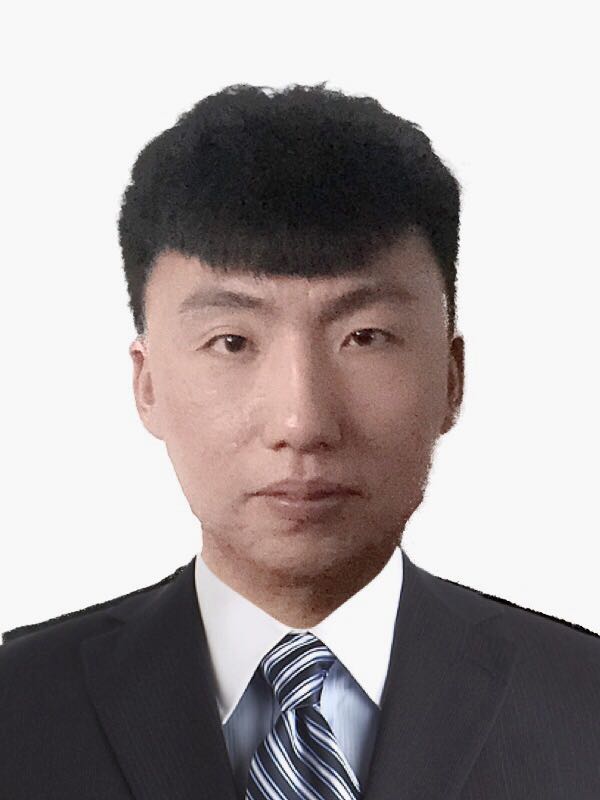}}]{Wen-Yu Shao}
	received the B.S degree in engineering mechanics from Taiyuan University of Science and Technology,Taiyuan, China, in 2020. He is currently pursuing the M.S. degree in electronic information at Taiyuan University of Science and Technology, Taiyuan, China. His current research interests include image fusion and deep learning.
\end{IEEEbiography}

\begin{IEEEbiography}[{\includegraphics[width=1in,height=1.25in,clip,keepaspectratio]{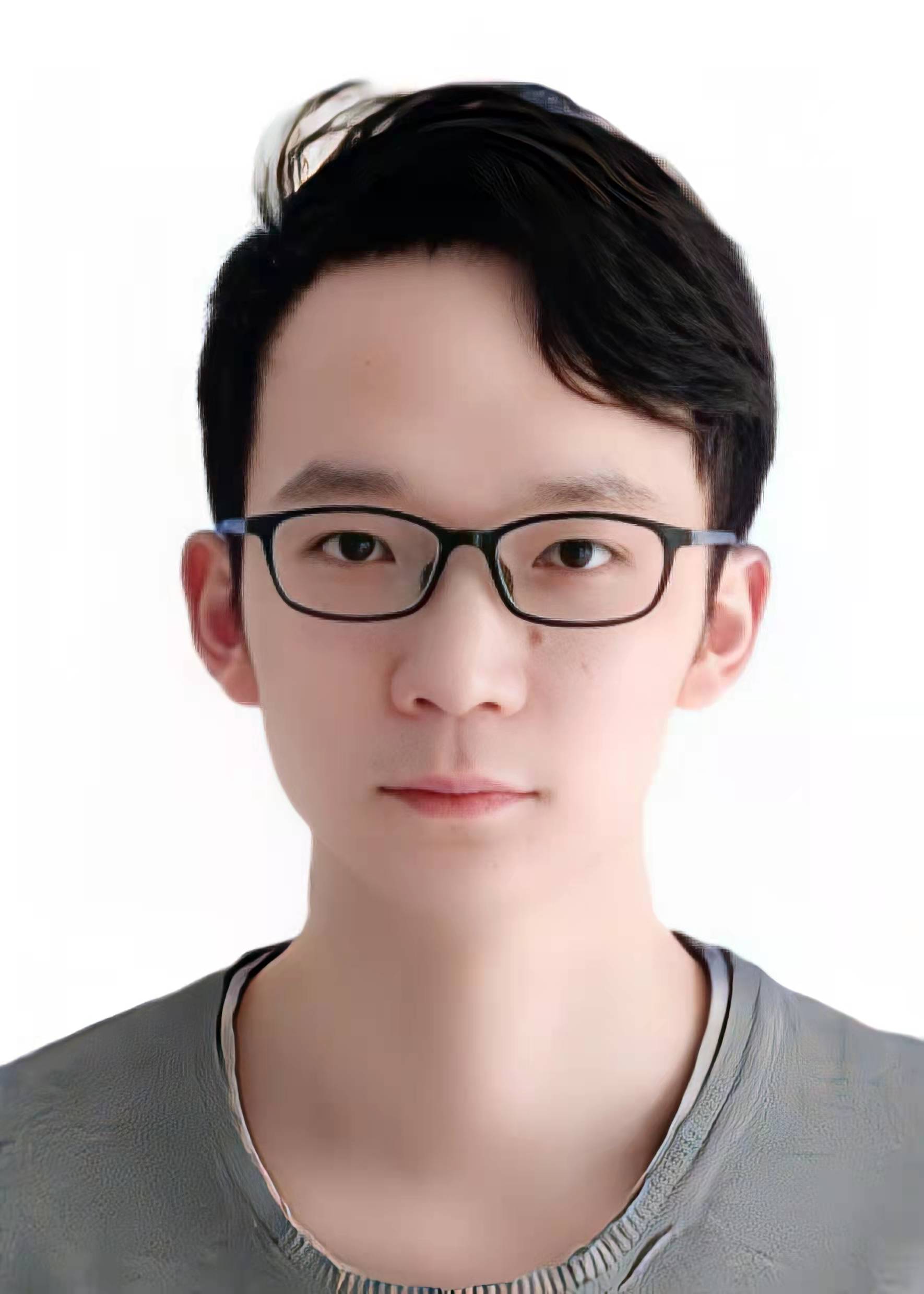}}]{Yan-Lin Chen}
	received the B.S degree in communication engineering from Hunan University of Technology, Zhuzhou, China, in 2019. He is currently pursuing the M.S. degree in optical engineering at Taiyuan University of Science and Technology, Taiyuan, China. His current research interests include image fusion and deep learning.
\end{IEEEbiography}

\begin{IEEEbiography}[{\includegraphics[width=1in,height=1.25in,clip,keepaspectratio]{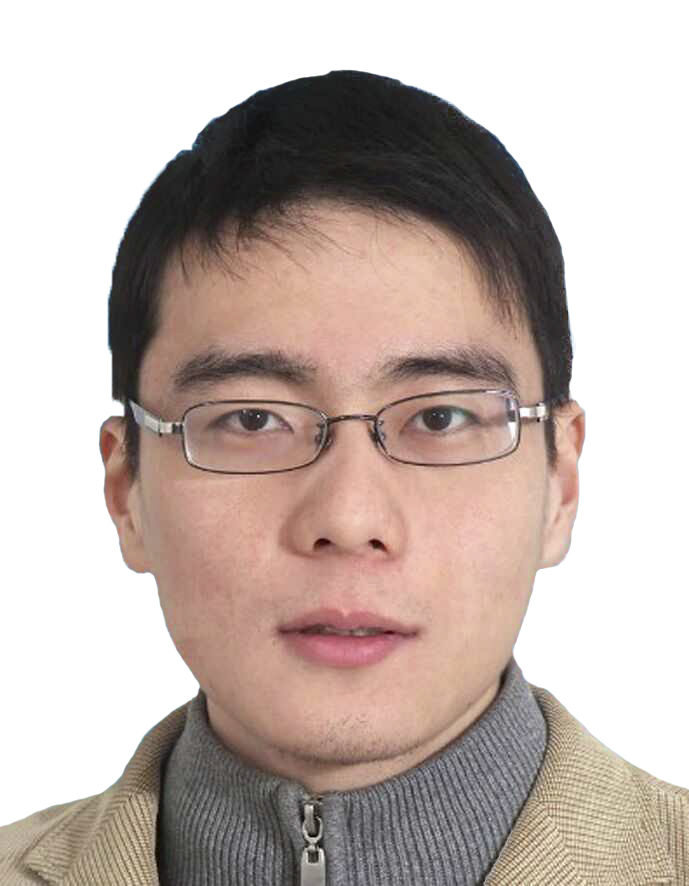}}]{Jia-Wei Xu}
    is with the Institute of Big Data and Information Technology, Wenzhou University, Wenzhou ,China. He is also with the College of Computer Science and Artificial Intelligence, Wenzhou University, Wenzhou, China from 2020. He was with School of Computing, Newcastle University from 2015 to 2019. He received Ph.D. degree in Eye-tracking lab, University of Lincoln, UK. His research interests include human factors in driving, such as driver eye movement, driver behaviors.
        
\end{IEEEbiography}

\begin{IEEEbiography}[{\includegraphics[width=1in,height=1.25in,clip,keepaspectratio]{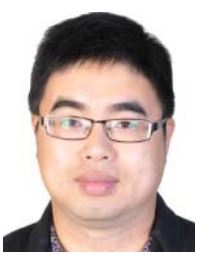}}]{Xiao-Qin Zhang} \textit{(Member, IEEE)} received the B.S degree in electronic information science and technology from Central South University, China, in 2005, and the Ph.D. degree in pattern recognition and intelligent system from the National Laboratory of Pattern Recognition, Institute of Automation, Chinese Academy of Sciences, China, in 2010. He is currently a Professor with Wenzhou University, China. His research interests include pattern recognition, computer vision, and machine learning.
\end{IEEEbiography}

%\newpage


%\vfill


\end{document}